%% file: main.tex
\documentclass[sigconf]{acmart}

\AtBeginDocument{%
  \providecommand\BibTeX{{%
    \normalfont B\kern-0.5em{\scshape i\kern-0.25em b}\kern-0.8em\TeX}}}

\setcopyright{acmlicensed}
\copyrightyear{2024} 
\acmYear{2024} 
\setcopyright{acmlicensed}\acmConference[KDD '24]{Proceedings of the 30th ACM SIGKDD Conference on Knowledge Discovery and Data Mining}{August 25--29, 2024}{Barcelona, Spain}
\acmBooktitle{Proceedings of the 30th ACM SIGKDD Conference on Knowledge Discovery and Data Mining (KDD '24), August 25--29, 2024, Barcelona, Spain}
\acmDOI{10.1145/3637528.3671586}
\acmISBN{979-8-4007-0490-1/24/08}

\usepackage{algorithm,algorithmic}
\usepackage{enumitem}

\usepackage{amsfonts}       % blackboard math symbols
\usepackage{nicefrac}       % compact symbols for 1/2, etc.
\usepackage{microtype}      % microtypography
\usepackage{xcolor}         % colors
\usepackage{utfsym}

\usepackage{subfigure}
\usepackage{xspace}

\usepackage{booktabs}       % professional-quality tables
\usepackage{MnSymbol}
\begin{document}

\title{SepsisLab: Early Sepsis Prediction with Uncertainty Quantification and Active Sensing}

\author{Changchang Yin}
\email{yin.731@osu.edu}
\affiliation{
  \institution{The Ohio State University} 
  \city{Columbus}
  \state{Ohio}
  \country{USA}
}

\author{Pin-Yu Chen}
\email{pin-yu.chen@ibm.com}
\affiliation{
  \institution{IBM Research} 
  \city{Yorktown Heights}
  \state{New York}
  \country{USA}
}

\author{Bingsheng Yao}
\email{b.yao@northeastern.edu}
\affiliation{
  \institution{Northestern University} 
  \city{Boston}
  \state{Massachusetts}
  \country{USA}
}

\author{Dakuo Wang}
\email{d.wang@northeastern.edu}
\affiliation{
  \institution{Northestern University} 
  \city{Boston}
  \state{Massachusetts}
  \country{USA}
}

\author{Jeffrey Caterino}
\email{jeffrey.caterino@osumc.edu}
\affiliation{
  \institution{The Ohio State University Wexner Medical Center} 
  \city{Columbus}
  \state{Ohio}
  \country{USA}
}

\author{Ping Zhang}
\email{zhang.10631@osu.edu}
\affiliation{
  \institution{The Ohio State University} 
  \city{Columbus}
  \state{Ohio}
  \country{USA}
}
\authornote{Corresponding Author}

\renewcommand{\shortauthors}{Changchang Yin et al.}

\newcommand{\CY}[1]{\textcolor{red}{(#1)}}
\newcommand{\PZ}[1]{\textcolor{blue}{(#1)}}
\newcommand{\system}{SepsisLab\xspace}
\input{abstract}







\maketitle 

\input{intro}
\input{relatedwork}

% \input{uncertainty}
\input{method}

\input{experiment}

\input{conclusion}
\input{acknowledgement}
\clearpage

\bibliographystyle{ACM-Reference-Format}
\bibliography{sample-base}
% \clearpage

\input{appendix}

\end{document}

%% file: abstract.tex
\begin{abstract}

Sepsis is the leading cause of in-hospital mortality in the USA. Early sepsis onset prediction and diagnosis could significantly improve the survival of sepsis patients. Existing predictive models are usually trained on high-quality data with few missing information, while missing values widely exist in real-world clinical scenarios (especially in the first hours of admissions to the hospital), which causes a significant decrease in accuracy and an increase in uncertainty for the predictive models. The common method to handle missing values is imputation, which replaces the unavailable variables with estimates from the observed data. The uncertainty of imputation results can be propagated to the sepsis prediction outputs, which have not been studied in existing works on either sepsis prediction or uncertainty quantification. In this study, we first define such propagated uncertainty as the variance of prediction output and then introduce uncertainty propagation methods to quantify the propagated uncertainty. Moreover, for the potential high-risk patients with low confidence due to limited observations, we propose a robust active sensing algorithm to increase confidence by actively recommending clinicians to observe the most informative variables. We validate the proposed models in both publicly available data (i.e., MIMIC-III and AmsterdamUMCdb) and proprietary data in The Ohio State University Wexner Medical Center (OSUWMC). The experimental results show that the propagated uncertainty is dominant at the beginning of admissions to hospitals and the proposed algorithm outperforms state-of-the-art active sensing methods. Finally, we implement a SepsisLab system for early sepsis prediction and active sensing based on our pre-trained models. Clinicians and potential sepsis patients can benefit from the system in early prediction and diagnosis of sepsis.

\end{abstract}

%% file: intro.tex
\section{Introduction}
Sepsis, defined as life-threatening organ dysfunction in response to infection, contributes to up to half of all hospital deaths and is associated with more than \$24 billion in annual costs in the United States \cite{liujama}. Existing studies \cite{liu2017timing} have shown that a sepsis patient may benefit from a 4\% higher chance of survival if they are diagnosed 1 hour earlier, so developing an early sepsis onset prediction system can significantly improve clinical outcomes.

% Existing works \cite{} usually use a pre-defined list of clinical variables (including demographics, vital signs, and lab tests) to make the sepsis prediction and presume all the variables in the list are present during the inference phase. \PZ{Really such a flow? We may start from existing predictive models are usually trained based on high-quality data with few missing information.} 
Existing machine-learning-based predictive models~\cite{islam2019prediction, reyna2019early,zhang2021interpretable,kamal2020interpretable} are usually trained on high-quality data with few missing information, while missing values widely exist in emergency department (ED) and emergency medical services (EMS) settings, which would cause most existing sepsis prediction models to suffer from performance decline and high uncertainty.
In addition, existing studies \cite{scott2018sensitivity,tsertsvadze2015community} have shown that for sepsis cases, most patients have already progressed into sepsis before the admissions to hospitals or during the first hours of admissions.
Thus it is critical to develop accurate sepsis prediction systems that can handle high missing-rate settings (e.g., cold-start setting with only several limited vital signs).

% Due to the success of deep learning algorithms, there is a growing interest from both academia and industry in the development of artificial intelligence (AI) to support clinical decision-making. Despite the yearly emergence of numerous studies boasting superior performance, their practical deployment in real-world clinical settings remain limited. A pivotal obstacle to their widespread adoption lies in the deficiency of uncertainty quantification, which raises concern when applying DL models to high-stake settings (e.g., early sepsis onset prediction).
% Missing data is an unavoidable consequence of tabular data collection in many domains. Nonetheless, most statistical studies and machine learning algorithms not only assume complete data, but cannot run on data sets with missing entries. 
A common method to handle missing variables is imputation, in which missing values are replaced by estimates from the observed data. To use the existing methods, we will need data imputations, which come with a new problem for the downstream sepsis prediction tasks: the uncertainty of imputation results can propagate to the sepsis prediction models.
Especially for deep learning models, a small perturbation in the input variables might cause a significant change in the predicted risk~\cite{madry2017towards,ren2020adversarial}. 
When the prediction models are sensitive to the highly uncertain input (i.e., imputed variable), the generated outputs are not reliable, so it is critical to quantify and reduce such kind of uncertainty. 
However, unlike epistemic uncertainty~\cite{swiler2009epistemic} and aleatory uncertainty~\cite{der2009aleatory}, the propagated uncertainty from the (imputed) input has not been investigated.  

\begin{figure}[]
    \centering
    \includegraphics[width=0.98\linewidth]{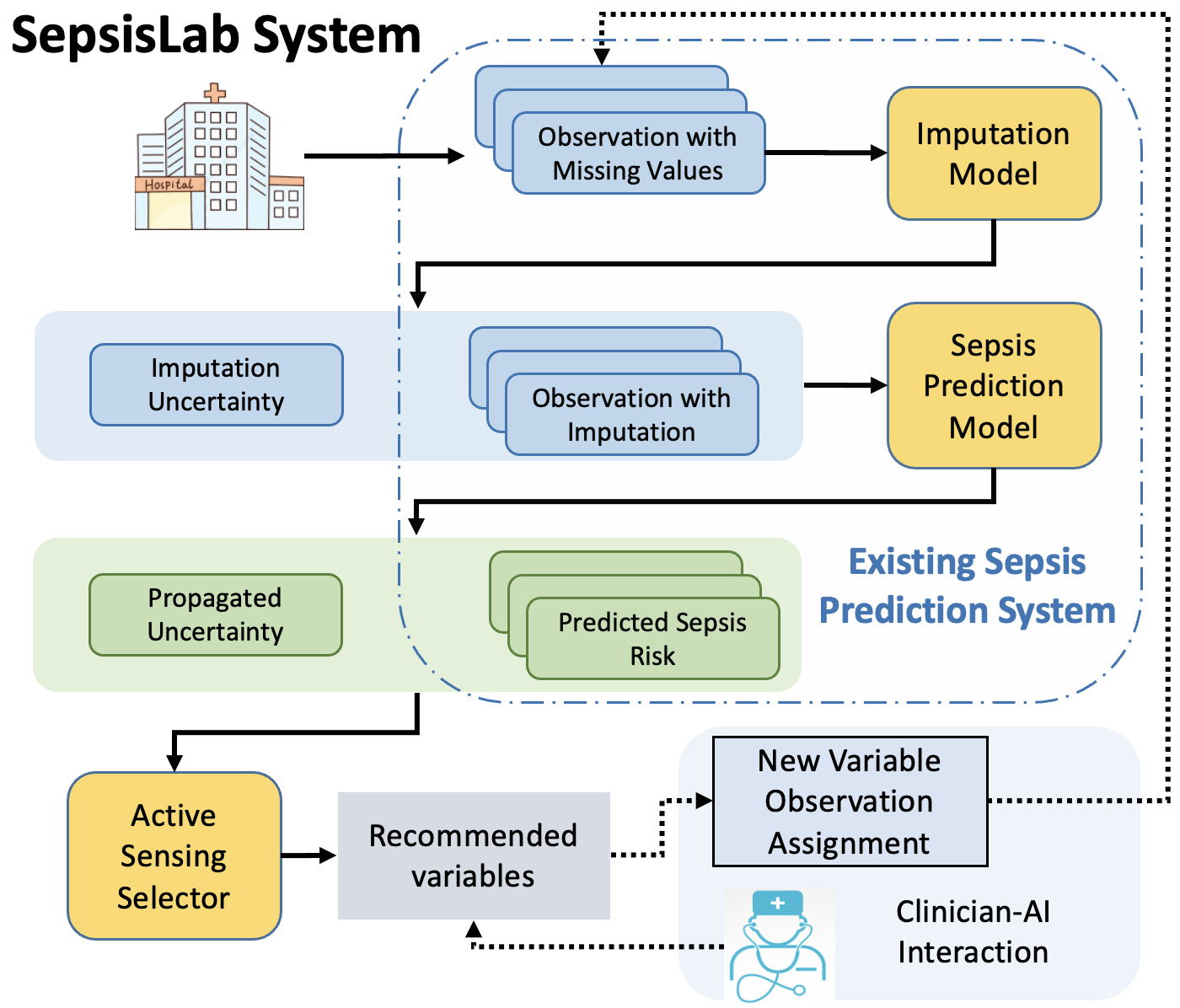}
    \vspace{-4mm}
    \caption{Workflow of \system system.}
    \label{fig:workflow}
    % \vspace{-4mm}
\end{figure}

In this study, we develop an early sepsis prediction system  \system that can quantify and reduce such kind of propagated uncertainty from missing value and imputation. 
\autoref{fig:workflow} displays the workflow of \system system.
% We first define the propagated uncertainty as the variance of the output with the consideration of the distribution of missing values.  
% We split the different-source uncertainties into two terms. The uncertainty from the weights can be modeled with various existing methods.
Given a patient's data with limited observations, we first adopt an imputation model to estimate the distribution (i.e., mean and standard deviation) of missing values. The standard deviation can be treated as the uncertainty of the imputed results.
Then we propose a time-aware sepsis prediction model to predict whether the patients will suffer from sepsis in the coming hours. The prediction model can generate sepsis risk and uncertainty simultaneously.
Given the estimated uncertainty, we further propose a robust active sensing algorithm to recommend clinicians observe the most informative lab test items that can maximally reduce the uncertainty for the potential high-risk patients.
% Note that most lab tests results will be available in less than 30~60 minutes\footnote{\label{test_time} \url{https://rb.gy/s4jiif}}, so \system can provide an updated sepsis risk with higher confidence soon. \PZ{Shall we delete this Note sentence here? It is clear enough now as I saw you have put this sentence in the experimental results part.}
% After the clinicians validate the recommended variables, new variable observation will be assigned and the additional observed lab test results will be added to the patient's EHR data. 
% \PZ{reviewers may ask how long the lab results come?} 
%We run the imputation and sepsis prediction model again to get the updated risk and uncertainty.
The active sensing module can significantly improve downstream sepsis prediction performance by providing more accurate observation and reducing the propagated uncertainty.

% \CY{Not sure whether it is better to put this paragraph into the last one. }
% Note that uncertainty propagation can be accurately calculated for linear functions while DL models are non-linear. We adopt adversarial learning to make the learned risk prediction function locally linear in the neighborhood of the mean value of the missing variables by minimizing the local linearity measure, which is helpful for accurately estimating the propagated uncertainty.
% Moreover, the local linearity can reduce the uncertainty caused by the observation bias and improve the risk prediction performance.  

% We design the active sensing experiments to validate the effectiveness of the propagated uncertainty qualification. Active sensing algorithm recommends clinicians requesting more variables (e.g., laboratory values from blood tests) for improving the confidence of the predictions. 
To demonstrate the effectiveness of the proposed models, we conduct experiments on real-world clinical datasets (including two publicly available datasets MIMIC-III~\cite{mimic} and AmsterdamUMCdb~\cite{amsterdamumcdb}, and proprietary data from The Ohio State University Wexner Medical Center (OSUWMC)). 
Experimental results show that the developed system can successfully work on both high- and low-missing-rate settings and achieve state-of-the-art sepsis prediction performance.
Finally, we develop a \system system for deployment to integrate into clinicians' workflow, which paves the way for human-AI collaboration and early intervention for sepsis management.

We summarize our contributions as follows:
\begin{itemize}[nosep,topsep=0pt,parsep=0pt,partopsep=0pt]
    \item We introduce propagated uncertainty to deep learning models, a new source of uncertainty different from widely studied aleatoric uncertainty and epistemic uncertainty
    \item We adopt uncertainty propagation to successfully qualify the propagated uncertainty, and the experimental results demonstrate the propagated uncertainty is dominant at the beginning of patients' admissions to hospital. 
    % and propose to reduce local linearity measure with adversarial training to more accurately compute and reduce the propagated uncertainty. 
    \item We propose a new active sensing framework RAS, which could effectively select variables to observe, and the experiments demonstrate the effectiveness of the proposed propagated uncertainty qualification method. 
    \item We design an interactive system SepsisLab\footnote{https://github.com/yinchangchang/SepsisLab} to make clinicians able to easily use and effectively interact with the models. 
\end{itemize}

% The paper is organized as follows: We briefly review the related work in \autoref{sec:related_work} and present the technical details for the imputation model, sepsis prediction model, and robust active sensing algorithm in \autoref{sec:method}. Then we describe the experimental setup in \autoref{sec:experiment} and display the experimental results in \autoref{sec:result}. We show how the implemented \system system for early sepsis prediction and active sensing based on our pre-trained models is deployed in \autoref{sec:deployment}. Finally, we concludes our work in \autoref{sec:conclusion}.  

%% file: relatedwork.tex
\begin{figure*}[!ht]
    \centering
    \vspace{4mm}
    \includegraphics[width=\textwidth]{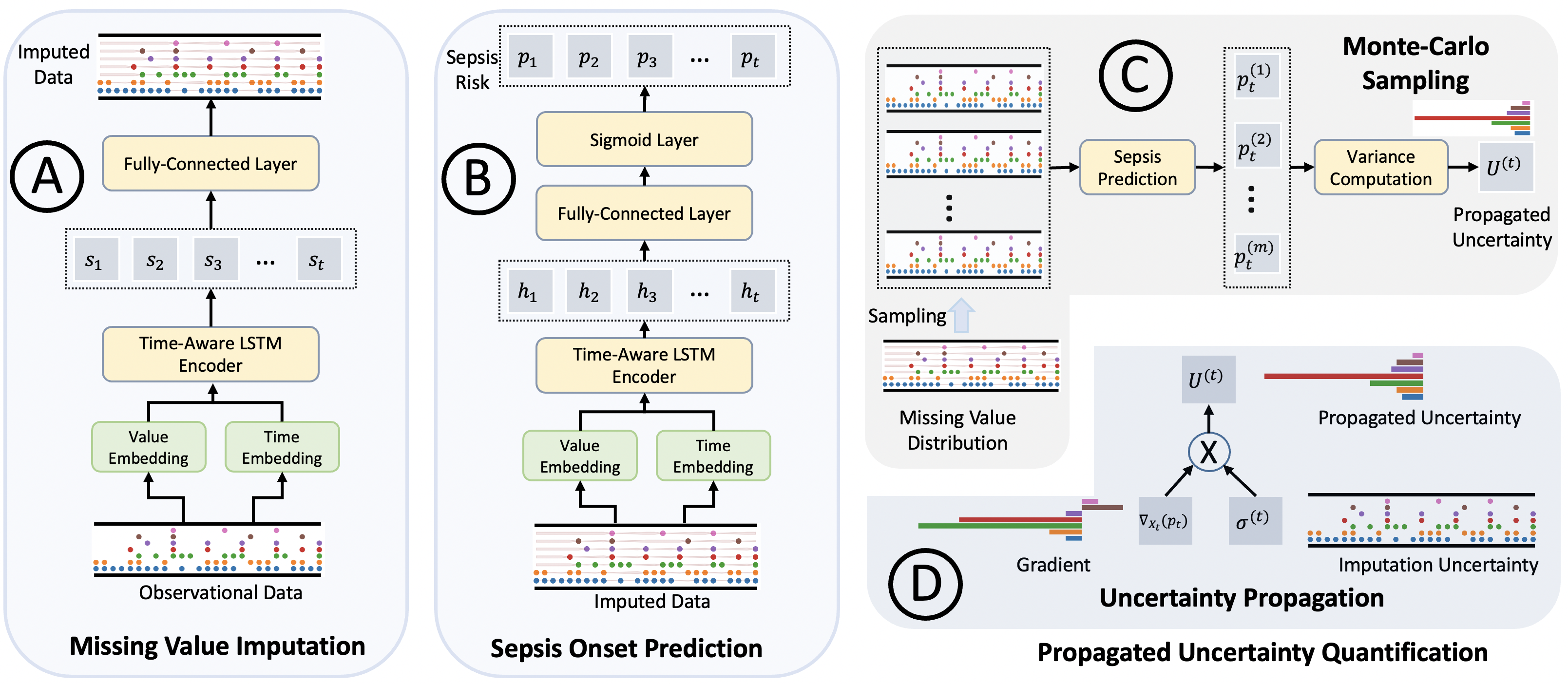}
    \caption{Model framework. (A) The imputation model takes observed variables and corresponding timestamps as input, and generates the distribution of missing values. (B) Sepsis prediction model produces the patients' sepsis onset risks with uncertainty based on the imputed data. (C) shows the uncertainty quantification method with Monte-Carlo sampling. (D) displays the uncertainty propagation method that can estimate propagated uncertainty by multiplying models' gradient over imputed variables and the imputation uncertainty.}
    \label{fig:model}
    % \vspace{-4mm}
\end{figure*}
\section{Related Work}
\label{sec:related_work}
In this section, we briefly review the existing studies related to sepsis prediction systems,  uncertainty qualification and active sensing. 

\subsection{Sepsis Prediction Systems} 

Sepsis is a heterogeneous clinical syndrome that is the leading cause of mortality in hospital intensive care units (ICUs) \cite{sepsis3,yin2020identifying}. Early prediction and diagnosis may allow for timely treatment and lead to more targeted clinical interventions. 
Screening tools have been used clinically to recognize sepsis, including qSOFA \cite{singer2016third}, MEWS \cite{subbe2001validation}, NEWS \cite{smith2013ability}, and SIRS \cite{bone1992definitions}. However, those tools were designed to screen existing symptoms as opposed to explicitly early predicting sepsis before its onset, and their efficacy in sepsis diagnosis is limited. 
% For example, prior studies show that qSOFA had low sensitivities in identifying sepsis in both pre-hospital and emergency department (ED) settings \cite{dorsett2017qsofa,usman2019comparison}
With recent advances, deep learning methods have shown great potential for accurate sepsis prediction \cite{islam2019prediction, reyna2019early,zhang2021interpretable,kamal2020interpretable}.
Although the methods achieved superior performance, they face a critical limitation: the models need to take the complete observation of a list of variables (including vital signs and lab tests), while lots of variables are missing in real-world data (especially in the first hours of admissions). Existing studies~\cite{zhang2021interpretable,kamal2020interpretable,islam2019prediction} usually impute the missing values before the prediction, which raises a new problem that the sepsis prediction models will heavily rely on the imputation methods. The imputation uncertainty would also be propagated to downstream prediction models. Thus it is necessary to quantify the propagated uncertainty, especially for high-stakes sepsis prediction tasks.

% Early sepsis decision-making needs to happen as soon as the patient arrives at ED --- a septic patient may benefit 4\% higher chance of survival if they are diagnosed 1-hour earlier~\cite{liu2017timing}, but 20.8\% of septic patients could have been diagnosed earlier\cite{neilson2023diagnostic}, because the physician often does not have as much information (e.g., WBC or C-Reactive~\cite{luo2022evaluating,van2006imputation}) about the patient as in other medical decision-making settings (e.g., an oncologist making a decision with many slides of scan images of a patient~\cite{pan2018synthesizing,chenyang2020joint}), thus they need to make a decision under \textbf{high uncertainty}.   

\subsection{Uncertainty Qualification}
Understanding what a model does not know is a critical part of many machine learning systems. Despite the superior performance deep learning models have achieved in various domain, they are usually over-confident about the predictions, which could limit their applications to real-world risk-sensitive settings (e.g., in healthcare). Uncertainty quantification methods play a pivotal role in reducing the impact of uncertainties during both optimization and decision making processes \cite{abdar2021review}. Existing uncertainty qualification work  \cite{kendall2017uncertainties,der2009aleatory,senge2014reliable,hora1996aleatory} has widely studied epistemic uncertainty and aleatoric uncertainty. However, most existing uncertainty qualification studies ignore an important uncertainty source: the uncertainty propagated from the uncertainty of input (e.g., widely existing missing values). In this study, we aim to investigate and reduce the propagated uncertainty. 

\subsection{Active Sensing}
Active sensing aims to improve the target tasks' performance by actively selecting most informative variables with the minimal cost. Yu et. al \cite{yu2009active} propose to select the informative variables based on mutual information and predictive variance. However, the model is based on Bayesian co-training framework, the prediction ability of which is not as good as deep neural networks when handling large-scale time serial data. 
Yoon et. al \cite{yoon2018deep} attempt to solve the active sensing problem by proposing an RNN-based model (i.e., Deep Sensing). 
The Deep Sensing framework involves learning 3 different networks: an interpolation network, a
prediction network and an error estimation network. Each network is separately optimized
for its own objective and then combined together after training to be used for active sensing.
Jarrett et. al \cite{jarrett2020inverse} propose an Inverse Active Sensing (IAS) to require negotiating (subjective) trade-off between accuracy, speediness, and cost of information. Yoon et. al \cite{yoon2019asac} propose an RL-based framework (Active Sensing using Actor-Critic models, ASAC) to directly optimize the predictive power after active sensing. Although the methods achieved superior performance in the target prediction tasks, they failed to measure the uncertainty of both missing values and model output risks, which limit their application in high-stakes clinical settings.

In this study, we aim to develop an accurate sepsis prediction system with propagated uncertainty quantification and incorporate active sensing algorithms to reduce the propagated uncertainty.

% \textbf{Local linearity with adversarial training.}Despite their superior performance, deep learning models are vulnerable to visually imperceptible but carefully chosen adversarial perturbations which cause neural networks to output incorrect predictions \cite{szegedy2013intriguing}. It is critical to overcome such vulnerability when applying deep learning models to risk-sensitive settings. Existing studies \cite{miyato2018virtual,qin2019adversarial,moosavi2019robustness} have shown that adversarial learning can make the learned function more smooth and locally linear. Moosavi-Dezfooli et. al \cite{moosavi2019robustness} find that adversarial training can lead to a significant decrease in the curvature of the loss surface with respect to inputs, leading to a drastically more “linear” behaviour of the network. They propose a new regularizer that directly minimizes curvature of the loss surface, and leads to adversarial robustness.  Miyato et. al \cite{miyato2018virtual} propose a virtual adversarial training to optimize a new measure of local smoothness of the conditional label distribution given input.  Qin et. al \cite{qin2019adversarial} propose to directly optimize the local linearity to make the model more robust. The studies' results show that the learned model can achieve better performance and become more  robust against adversarial, norm-bounded perturbations. In this study, we also introduce the adversarial training to make the model more robust, which is also helpful for accurate computation of uncertainty propagation. 

%% file: method.tex
\section{Methodology}
\label{sec:method}
In this section, we present the proposed sepsis prediction system \system, including a missing value imputation model, an early sepsis prediction model, and an active sensing algorithm. 

% first present our sepsis onset prediction model.
% We briefly introduce various kinds of uncertainty and then define propagated uncertainty. Then we introduce an uncertainty propagation method to quantify the uncertainty. Moreover, we present the proposed robust active sensing framework and the interactive visualization system. 

\subsection{Notation and Problem Statement}   
In this study, we aim to predict sepsis onset with limited clinical variables observed.  
We consider the following setup. 
A patient has a sequence of clinical variables (i.e., lab test data and vital sign data) with timestamps. 
Let $Z \in \{R \cup *\}^{n \times k}$ denote the observations of variables, where $*$ represents missing values, $n$ denotes the number of collections of observations and $k$ denotes the number of unique clinical variables. $T \in R^n$ denotes the observation timestamps.
$Y \in \{0, 1\}^n$ denotes the ground truth of whether the patient will progress to sepsis in the coming hours. Following \cite{zhang2021interpretable,kamal2020interpretable}, we set the prediction window as 4 hours.
Due to the existence of missing values, we impute the missing values first and use $X\in R^{n \times k}$ to denote the imputed results.

Given a loss function $\mathcal{L}$ and a distribution over pairs ($X$, Y), the goal is to find a function $f$ that minimize the expected loss:
\begin{equation}
    f^* = \arg \min_f E [\mathcal{L}(f(X), Y)]
    % \vspace{-4mm}
\end{equation}

We list the important notations in \autoref{tab:notation}.

\begin{table}[]
    \centering
    \caption{Basic Notations.}
    \begin{tabular}{cl}
        \toprule
        Notation & Description \\
        \midrule
        $Z$ & Observed variables with missing values.\\
        $X$ & Imputed variables. \\
        $Y$ & Labels for sepsis prediction. \\
        $p_i$ & Predicted sepsis risk at $i^{th}$ collection. \\
        $T$ & Timestamps for observations.\\
        $M$ & Masking indicator for imputation training. \\
        $n$ & The number of collections of variables.\\
        $k$ & The number of unique variables.\\
        $e^t_i$ & Time embedding vector for $i^{th}$ collection. \\
        $e_i$ & The embedding for $i^{th}$ collection. \\
        $\mu$ & Mean of missing values. \\
        $\sigma$ & Standard deviation of missing values. \\
        $w_*, b_*$   & Learnable parameters. \\
        $s_i$ & Hidden state of imputation model. \\
        $h_i$ & Hidden state of the sepsis prediction model. \\ 
        $U$ & Computed uncertainty. \\
        $U_x$ & Propagated uncertainty. \\
        $U_w$ & Epistemic uncertainty. \\
        $\rho{ij}$ & Correlation between $i^{th}$ and $j^{th} variable$. \\
        $lr$ & Learning rate. \\
         \bottomrule
    \end{tabular}
    \label{tab:notation}
\end{table}

\subsection{Missing Value Imputation}
We assume the missing values follow the Gaussian distributions and impute the missing values by estimating the distribution of variables (i.e., the mean and covariance). 
\autoref{fig:model}(A) shows the framework of our imputation model.

Following~\cite{sciencesubtyping}, we first use mean-imputation to preprocess the observational data $Z$ and send the embedding of $Z$ to LSTM to model the patient's health states. 

\textbf{Embedding layer.} 
In the $i^{th}$ collection, we have  observational values $Z_i$, observation time $T_i$. We use a fully connected layer to embed the observed variable in the collection:
\begin{equation}
    e_i = w_e [Z_i; e^t_i] + b_e,
    \label{eq:embedding}
\end{equation}
where $[\bullet;\bullet]$ denotes concatenation operation. $w_e \in R^{(k+2d) \times d}$ and $b_e \in R^d$ are learnable variables. $e^t_i \in R^{2d}$ denotes the time embedding and is computed as follows:
\begin{equation}
\label{eq:time_embedding}
\begin{split}
e^t_{i,j} = sin(\frac{T_i*j}{T_{max}*d}), 
e^t_{i,d+j} = cos(\frac{T_i*j}{T_{max}*d}),
\end{split}
\end{equation}
where $0 \leq j < d$, and $T_{max}$ denotes the max value of $T$.

\textbf{Time-aware LSTM encoder.}
Given the embedding vectors $[e_1, e_2, ..., e_n]$, we use LSTM to model the patients' states:
\begin{equation}
    s_1, s_2, ..., s_T = LSTM(e_1, e_2, ..., e_n)
\end{equation}

\textbf{Missing value distribution estimation.}
% We assume the missing values follow the Gaussian distribution and the correlation between variables is invariant in the data. For example, the correlation $\rho_{ij}$ between $i^{th}$ and $j^{th}$ variable can be directly computed from the observed data, so we focus on the estimation of mean $\mu$ and standard deviation $\sigma$ for the missing values.
A fully connected layers is used to generate the parameters of the missing value distribution: 
\begin{equation}
    \mu_{i} = w_\mu s_i + b_\mu, \qquad
    \sigma_{i} = ReLU(w_\sigma s_i + b_\sigma),
    \label{eq:distribution}
\end{equation}
where $w_\mu, w_\sigma\in R^k$ and $b_\mu, b_\sigma \in R$ are learnable variables.

We train the imputation model 
% $f_{imp}$ 
with the mean square error loss function:
% \vspace{-3mm}
\begin{equation}
    \label{eq:mu}
    \mathcal{L}_{imp}(Z, M, \mu) = \sum_{i=1}^n \sum_{j=1}^k M_{i,j} (\mu_{i,j} - Z_{i,j})^2,
    % \vspace{-3mm}
\end{equation}
where $M \in \{0,1\}^{n \times k}$ denotes the indices of masked variables.
$M_{i, j}$ is 1 if the $j^{th}$ variable in $i^{th}$ collection is observed and masked; otherwise, 0.
Replacing the missed values $*$ with the estimates $\mu$, the observed variables $Z$ become $X \in R^{T \times n}$.

After the imputation model is well-trained with \autoref{eq:mu}, we further learn to estimate the standard deviation $\sigma$ by finetuning $w_\sigma$ and $b_\sigma$ and fixing other parameters. We minimize the following loglikelihood loss:
% \vspace{-2mm}
\begin{equation}
    \label{eq:sigma}
    \mathcal{L}_\sigma(Z, M, \mu, \sigma) = \sum_{i=1}^n \sum_{j=1}^k  M_{i,j} [\frac{(\mu_{i,j} - Z_{i,j})^2}{2 \sigma_{i,j}^2} + \frac{\sigma_{i,j}^2}{2}]
    % \vspace{-2mm}
\end{equation}
 
% Then we will continue to learn the sepsis onset prediction function with the updated input: $f^* = \arg\min_f E [\mathcal{L}(f(D, \Tilde{X}), Y)]$. 

\subsection{Sepsis Prediction Model}

With the imputed results to replace the missing values, we continue to predict whether the patients will suffer from sepsis in the coming hours. 
The framework of sepsis prediction model is shown in
\autoref{fig:model}(B). 

Similar to \autoref{eq:embedding} in the imputation model, 
we use the same embedding layers in the imputation model.
\begin{equation}
    e'_i = w_e [X_i; e^t_i] + b_e,
    \label{eq:embedding}
\end{equation}
where the time embedding $e^t_i$ is the same as in \autoref{eq:time_embedding}.

Then we use LSTM~\cite{lstm} to model the patient's health states.
A fully connected layer and a Sigmoid layer is followed to generate the sepsis risks:
\begin{equation}
    p_i = Sigmoid(w_s h_i + b_s), \text{ where } t = 1, 2, ..., T
    \label{eq:risk_prediction}
\end{equation}
\begin{equation}
    h_1, h_2, ..., h_n = LSTM(e'_1, e'_2, ..., e'_n),
\end{equation}
where $w_s \in R^d$ and $b_s \in R$ are learnable parameters.

The model is trained by minimizing the binary cross-entropy loss:
\begin{equation}
    \label{eq:cls_loss}
    \mathcal{L}_{cls}(p, Y)  = \frac{1}{n}\sum_{i=1}^n - y_i\log(p_i)  - (1-y_i) \log(1-p_i)
\end{equation}

\iffalse
Given the loss function $\mathcal{L}$ and a distribution over pairs $(Z, Y)$, the goal is to find a function $f$ to minimize the expected loss: 
\begin{equation}
    f^* = \arg\min_f E [\mathcal{L}(f(g(Z)), Y)],
\end{equation}
where $g(Z)$ denotes the imputation function. 
The function $f$ can be a pipeline that combines an imputation method with a prediction method. 
\fi

\subsection{Sources of Uncertainty}
% \CY{The paragraph will be re-written.}
% Although deep learning models achieved superior performance, they also suffer from overconfidence~\cite{}. 
% Such kind of deep learning prediction models is not reliable due to a lack of uncertainty quantification.
% In this subsection, we first investigate various sources of uncertainties and propose two kinds of methods to quantify the propagated uncertainty. 
 
% \subsection{Problem Formulation}
% \subsubsection{Sources of Uncertainty} 

% Although deep-learning-based models can achieve superior performance, 

% The generated sepsis risk in \autoref{eq:risk_prediction} could be uncertain, which limits the models' application to real-world clinical settings. 

% When patients arrive at hospitals, most variables (e.g., lab tests) are not available, which could cause the sepsis prediction models to suffer from high uncertainty.

When applying deep learning methods to high-stakes sepsis prediction tasks, the lack of uncertainty quantification will make the models less reliable. 
In this subsection, we investigate two main sources of uncertainty.
% : model parameter uncertainty and missing value uncertainty.
% In this study, we focus on the uncertainty sources that could directly guide the active sensing algorithms.   

\textbf{Uncertainty from the model parameters.}
Existing uncertainty qualification work  \cite{kendall2017uncertainties,der2009aleatory,senge2014reliable,hora1996aleatory} has widely studied epistemic uncertainty, which accounts for uncertainty in the model parameters, especially for the huge amount of parameters in deep learning models.
% The learned model parameters could directly affect the qualification of uncertainty from the unobserved variables and observation bias. It is necessary to take into account the epistemic uncertainty in sepsis onset prediction tasks. 
Following \cite{kendall2017uncertainties}, we use drop-out during the test phase and run the inference many times to quantify such kind of uncertainty. 

\textbf{Uncertainty from missing values.}  
Superior risk prediction models in the healthcare domain heavily rely on high-quality complete input. However,  missing values (e.g., vital signs and lab test results) widely exist in real-world clinical settings. 
% It would be harmful for patients to frequently observe some variables (e.g., frequent blood tests), so we have to predict the clinical risks based on the available but limited observations.
Most risk prediction methods~\cite{islam2019prediction, reyna2019early,zhang2021interpretable,kamal2020interpretable} first impute the missing values and then make predictions based on the imputed values. The accuracy of the imputation methods can directly affect the performance of the predicted sepsis risks.  The uncertainty from the imputation results can be directly propagated to downstream sepsis prediction models.

% \textbf{Uncertainty from observation bias.} The observed clinical variables, including vital signs and lab tests, might have biases or even errors due to incorrect positioning, poor communication, poorly designed processes, and so on~\cite{plebani2006errors,maslove2016errors,mrazek2020errors}. The observation bias or errors could also propagate to the prediction output. Moreover, recent works \cite{finlayson2019adversarial,ma2021understanding} also show that medical deep learning methods can be fooled by adversarial attacks with small imperceptible perturbations. This raises safety concerns about the deployment of these methods in clinical settings. It would be better to consider the observation bias when estimating the uncertainty and make the deep learning methods more robust in such risk-sensitive clinical settings.   

% would make the uncertainty estimation and risk prediction models less accurate and robust, which should be considered in such risk-sensitive clinical settings. 

% focus on the three kinds of uncertainty related to active sensing tasks: epistemic uncertainty from the learned model parameters and two kinds of propagated uncertainty from the missing values and the observation bias.

\subsection{Uncertainty Definition}  

We use the variance of prediction models' output to define the two kinds of uncertainty mentioned above. 
Patients' data $X$ contains a sequence of collections of variables. We can use all the observations until the current collections to make predictions.
When applying active sensing algorithms to reduce the propagated uncertainty with additional observations, we can only request the variables in the current collection.

In the active sensing task, we only focus on the uncertainty related to the latest collection.
In the following subsections, for simplicity, at a given time $T_i$, we use $x$ to represent the $i^{th}$ observation (i.e., $X_i$), and use $f_w(x)$ rather than $f_w(X)$ to denote the predicted risk, where $w$ means all the learnable parameters in the sepsis prediction model.

% For simplicity, we use $f_w(x)$ to represent $f_w(X)$ when estimating propagated uncertainty, where $x$ denotes the last collection of variables in $X$.

We assume the input variables $x \in R^k$
and model parameters $w$ follow Gaussian distributions $\mathcal{N}(\mu_x, \sigma_x)$
and $\mathcal{N}(\mu_w, \sigma_w)$. 
$\mu_x \in R^k$ and $\sigma_x \in R^{k}$ can be estimated with \autoref{eq:distribution}. 
Let $y$ denote the sepsis prediction label for the patient at current time. 
 
Following existing studies~\cite{kendall2017uncertainties}, we define the uncertainty of predicted risk as the variance of model outcomes:
 \begin{equation}
    \label{eq:uncertainty_definition}
    \begin{aligned} 
        & U = \int_w \int_x   (f_w(x) - \mu_y)^2   \rho(x)  dx   \rho(w)dw = U_x + U_w
        % & \qquad  \qquad + \int_w \int_x  2(f_w(x) - \mu_{y_w})(\mu_{y_w} - \mu_y) \rho(x)  dx   \rho(w)dw \\
        % & \qquad  \quad=   \int_x  (f_w(x) - \mu_{y_w})^2 \rho(x)  dx +  \int_w  (\mu_{y_w} - \mu_y)^2 \rho(w) dw  \\
 \end{aligned}
\end{equation}
\begin{equation*} 
   \begin{aligned}
        \text{where } & U_x  = \int_w \int_x   (f_w(x) - \mu_{y_w})^2  \rho(x)  dx   \rho(w)dw, \\
        &  U_w = \int_w  (\mu_{y_w} - \mu_y)^2 \rho(w)dw, \\
        & \mu_{y_w} = \int_x  f_w(x )  \rho(x) dx, \\
        & \mu_y = \int_w \int_x   f_w(x )  \rho(x)  \rho(w) dx dw,
    \end{aligned}
\end{equation*}

\noindent
where $\rho(\bullet)$ denotes the density function.

% Active sensing algorithms select the input variables that contribute most uncertainty of the model outcome. 

We split the uncertainty into two terms. The second term $U_w$ is caused by the model uncertainty from the model parameters, so we just focus on the first term $U_x$ when actively selecting unobserved variables. 

When the model parameter $w$ is fixed, we can estimate the propagated uncertainty as: 

\begin{equation}
    \label{eq:propagated_uncertainty}
    U^{(w)}_x  =  \int_x   (f_w(x) - \mu_{y_w})^2  \rho(x)  dx 
\end{equation}

% With the definition of the two kinds of uncertainty, we continue to estimate them in following subsections.

\subsection{Propagated Uncertainty Quantification}

\subsubsection{Propagated Uncertainty for Linear Target Prediction}
When the sepsis risk prediction function is a linear function, $f_w(x)=\sum_j w_j x_j$, following \cite{kirchner2001data}, we compute the uncertainty in \autoref{eq:propagated_uncertainty} as:
\begin{equation} 
    \label{eq:uncertainty_for_linear_function}
    U^{(w)}_x = \sum_i w_i^2 \sigma_{x_i}^2 + \sum_i \sum_{j\neq i} w_i w_j \rho_{ij} \sigma_{x_i} \sigma_{x_j},
\end{equation} 
where $\rho_{ij}$ denotes the correlation between $i^{th}$ and $j^{th}$ variable.
It is easy to compute the propagated uncertainty for linear function based on \autoref{eq:uncertainty_for_linear_function} for linear function. The calculation details for \autoref{eq:uncertainty_for_linear_function} can be found in \autoref{sec:linear_uncertainty} in supplementary materials.

The propagated uncertainty reduction after observing $i^{th}$ variable is:
\begin{equation}
    \label{eq:uncertainty_reduction_linear}
    U^{(w)}_{x}(i) = w_i^2 \sigma_{x_i}^2 + \sum_{i\neq j} w_i w_j \rho_{ij} \sigma_{x_i}
\end{equation}

\begin{algorithm}[t]
\caption{Adversarial Training} 
\label{alg:AT}
\leftline{\textbf{Input}: observations $X$, missing value distribution $\mu_x$, $\sigma_x$, }
$\qquad $ outcome $Y$, step size $s_{adv}$, step $n_{adv}$, learning rate $lr$;
% \leftline{\textbf{Output}: predicted risk $f_w(x)$;}

\begin{algorithmic}[1] %[1] enables line numbers
\REPEAT
\STATE Sample a batch of patients' data, $x$, $\sigma_x$, $y$;
\STATE Initialize the perturbation $\delta$ with Gaussian distribution $N(0, \sigma_x)$ and constraint $-2\sigma_x < \delta < 2\sigma_x$;
\STATE Compute $f_w(x)$ and the first order gradient $\nabla_x$;
\FOR{$i=1, ..., n_{adv}$}
\STATE Calculate $g = \nabla_\delta g(\delta, x)$
\STATE Update $\delta = \delta + s_{adv} \times g$
\ENDFOR
\STATE Calculate loss $L$ in \autoref{eq:weighted_loss} and gradient $\nabla_w L$;
\STATE Update $w = w - lr \times \nabla_w L$;
\UNTIL{Convergence.}
\end{algorithmic}
\end{algorithm}

\subsubsection{Propagated Uncertainty for Non-Linear Target Prediction} 
For the non-linear sepsis prediction function, we use the Taylor expansion as approximate function:
\begin{equation}
    \Tilde{f}_w(x + \delta) = f_w(x)   + \delta^T \nabla_xf_w(x)
\end{equation}

We can use the uncertain propagation in \autoref{eq:uncertainty_for_linear_function} as the approximation of the uncertainty of non-linear function $f_w$. 
However, the propagated uncertainty estimation for non-linear functions are biased on account of using a truncated series expansion. The extent of this bias depends on the nature of the function. 

The absolute difference between the two values $\Tilde{f}_w(x + \delta)$ and $f_w(x+\delta)$ is:
\begin{equation}
    \label{eq:residual}
    g(\delta, x) = |f_w(x+\delta) - f_w(x)   + \delta^T \nabla_xf_w(x)|
\end{equation}

When $g(\delta, x)$ is small enough in the neighborhood near $\mu_x$ (i.e., $f_w$ is locally linear), the propagated uncertainty in \autoref{eq:uncertainty_for_linear_function} is still accurate and able to guide the active sensing.

\subsection{Robust Active Sensing}

\subsubsection{Adversarial Training for Local Linearity}
Existing studies \cite{qin2019adversarial,moosavi2019robustness} have shown that adversarial training can encourage the local linearity of the learned functions.
In this study, we adopt adversarial training to make the target prediction function locally linear in a neighborhood near the mean value of input $x$. 
\begin{equation}
    \mathcal{L}_{adv} = \min_w \max_\delta g(\delta, \mu_x) \text{, where } - 2\sigma_x < \delta < 2\sigma_x
\end{equation}

The risk prediction model is trained with a weighted sum of classification loss and adversarial loss:
\begin{equation}
    \label{eq:weighted_loss}
    \mathcal{L} = \alpha \mathcal{L}_{cls} + (1 - \alpha) \mathcal{L}_{adv},
\end{equation}
where $0 < \alpha < 1$ is a hyper-parameter.

We consider the quantity:
\begin{equation}
    \label{eq:max_error}
    \gamma(\sigma, x) = \max_{-2\sigma \le \delta \le 2\sigma} |f_w(x + \delta) - f_w(x) - \delta^T\nabla_x f_w(x)|
\end{equation}
to be a measure of how linear the surface is within a neighborhood near $x$. We call this quantity the
local linearity measure. 
The missing variables follow Gaussian distribution, so $\delta$ lies within two standard deviations with more than probability 95\%.
The uncertainty estimation error would be less than $\gamma(\sigma_x, x)$ with probability more than 95\%. 

Algorithm \autoref{alg:AT} describes the training process
of the sepsis prediction model.

\subsubsection{Active Sensing}
The approximate uncertainty of the risk prediction outcome is defined as:
\begin{equation}    \label{eq:uncertainty_for_nonlinear_function}
    U^{(w)}_x = \sum_i \nabla_{x_i}^2 \sigma_{x_i}^2 + \sum_i \sum_{j\neq i} \nabla_{x_i} \nabla_{x_j} \rho_{ij} \sigma_{x_i} \sigma_{x_j}
\end{equation}

The propagated uncertainty reduction after observing $i^{th}$ variable is:
\begin{equation}
    \label{eq:uncertainty_reduction}
    U^{(w)}_i = \nabla_{x_i}^2 \sigma_{x_i}^2 + \sum_{j\neq i} \nabla_{x_i} \nabla_{x_j} \rho_{ij} \sigma_{x_i} \sigma_{x_j}
\end{equation}

Considering the distribution of $w$, we use Monte-Carlo dropout to sample model parameters and use the average uncertainty of $U^{(w)}_x(i)$ to approximately compute $U_x(i)$:
\begin{equation}
    U_x(i) = \int_w U^{(w)}_x (i) \rho(w) dw
\end{equation}
 
We can select the unobserved variables based on the maximal uncertainty criterion. 
\begin{equation} 
    i^* = \arg \max_i U_x(i),
\end{equation}
where $i^*$ is the best variable to observe. 
\autoref{fig:model}(D) shows the workflow of propagated uncertainty quantification and active sensing methods.

%% file: experiment.tex
\section{Experiment Setup}
 
\label{sec:experiment}

To demonstrate the effectiveness of the proposed method, we conducted experiments on real-world datasets.
% two publicly available real-world datasets MIMIC-III \cite{mimic} and AmsterdamUMCdb \cite{amsterdamumcdb} and one private dataset extracted from OSUWMC.

\subsection{Datasets}
\textbf{Datasets.}
We validate our system on two publicly available datasets 
( MIMIC-III\footnote{\url{https://mimic.physionet.org/}} and AmsterdamUMCdb\footnote{\url{https://amsterdammedicaldatascience.nl}}) and one proprietary dataset extracted from OSUWMC\footnote{https://wexnermedical.osu.edu/}.
We first extracted all the sepsis patients with sepsis-3 criteria~\cite{sepsis3} in the datasets. For each sepsis patient, we select 1 control patient with the same demographics (i.e., age and gender).
We extracted 26 vital signs and lab tests from the datasets.  A detailed list of clinical variables can be found in supplementary materials.
The statistics of the three datasets are displayed in \autoref{tab:stati}.

 \noindent
\textbf{Variables Used for Sepsis Prediction.}
Following \cite{yin2020identifying}, we use following variables to model sepsis patients' health states:
 heart rate, Respratory, Temperature, Spo2, SysBP, DiasBP, MeanBP, Glucose, Bicarbonate, WBC, Bands, C-Reactive, BUN, GCS, Urineoutput, Creatinine, Platelet, Sodium, Hemoglobin, Chloride, Lactate, INR, PTT, Magnesium, Aniongap, Hematocrit, PT.

 The first 8 variables are immediately available vital signs. The missing rates of the variables can be found in \autoref{tab:all-variables} in \autoref{sec:missing-rate}.

% The statistics of extracted data from MIMIC-III and AmsterdamUMCdb are displayed in the supplementary materials. 

\begin{table} [!t]
\centering
\caption{Statistics of MIMIC-III and AmsterdamUMCdb} 
% \vspace{-3mm}
\label{tab:stati} 
\begin{tabular}{cccc}
\toprule
 & MIMIC & AmsterdamUMCdb  & OSUWMC\\
 \midrule
 \#. of patients & 21,686 & 6,560 & 85,181\\
 \#. of male & 11,862 & 3,412& 41,710\\
 \#. of female & 9,824 & 3,148& 43,471\\
 Age (mean $\pm$ std) & 60.7 $\pm$ 11.6 & 62.1 $\pm$ 12.3 & 59.3 $\pm$ 16.1 \\
 Missing rate & 65\% & 68\% & 75\%\\
 Sepsis rate & 32\% & 35\% & 29\%\\
 \bottomrule
\end{tabular} 
% \vspace{-3mm}
\end{table}

\subsection{Setup}
% We conducted the experiments in two settings as follows:

% \textbf{Observation.} In this setting, we only use all the observational data to quantify the different kinds of uncertainties. 

\begin{figure}
    \centering
    \includegraphics[width=\linewidth]{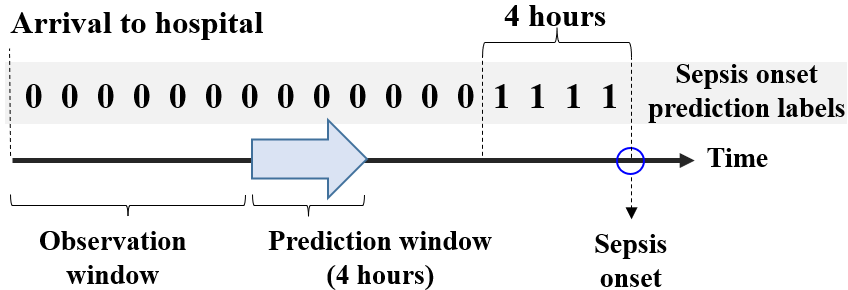}
    % \vspace{-3mm}
    \caption{Settings of sepsis onset prediction.
    % We predict whether patients will suffer from sepsis in 4 hours. 
    }
    % \vspace{-6mm}
    \label{fig:setting}
\end{figure}

% \textbf{Active sensing.}
We mimic the cold-start environment where only vital signs are immediately available, while all the lab tests can be observed after the assignment. 
\autoref{fig:setting} displays the setting of the experiments. After the patients arrive at the hospital, we start to predict whether the patients will suffer from sepsis in 4 hours. We run the prediction process hourly until the patients have been diagnosed with sepsis or discharged. 
When the model's output has a high uncertainty due to the limited observations, the active sensing algorithms can select the missing lab tests to observe. 
Based on the lab testing turn-around times policy of OSUWMC, most lab results will be available in less than 30\textasciitilde60 min\footnote{\label{test_time} \url{https://rb.gy/s4jiif}} (or even sooner for sepsis patients with high priority), so the observation results for the selected lab items can be used in the same hour to update the predicted sepsis risk.
Note that when active sensing algorithms select some variables that are not collected at the corresponding time, we use the estimates from other observed variables as the active observation results.

% \vspace{-3mm}

\subsection{Methods for Comparison}
We compare the proposed model with following methods:
\begin{itemize}[leftmargin=*]
    \item \textbf{Random sensing}: We randomly select the masked values to observe for random sensing.
    \item \textbf{Active sensing MI} \cite{yu2009active}: The method selects the most informative variables based on the mutual information. 
    % \item \textbf{Active learning:} Active learning select the most informative samples to label, which is similar to our setting. When the model is uncertain about the predictions, we request more variables to observe based on the mutual information.
    \item \textbf{Virtual adversarial training (VAT)} \cite{miyato2018virtual}: VAT  proposes to make the learned function locally linear with local a smoothness regularization method. Then we use the same variable selection method as ours to select missing values. 
    \item \textbf{Monte Carlo sampling:} Existing studies \cite{kendall2017uncertainties} use Monte-Carlo dropout to measure the epistemic uncertainty. Similarly, we use Monte-Carlo sampling to estimate the propagated uncertainty by sampling the values of the unobserved variables based on the Gaussian distribution and select the variable with maximal variance in generated output, as \autoref{fig:model}(C) shows.
    \item \textbf{Robust active sensing (RAS)}: RAS is the proposed method. To demonstrate the effectiveness of the adversarial training, we implement three versions of RAS. RAS is the main version. RAS$^L$ uses the linear constraint to make the learned function locally linear. RAS$^N$ means the model is only trained by minimizing the classification loss in \autoref{eq:cls_loss} without any linearity constraint.
\end{itemize}

For fair comparison to the baselines, all active sensing algorithms use the same deep-learning sepsis prediction model backbone. Our previous works~\cite{yin2020identifying,zhang2021interpretable} have shown that LSTM can successfully model the time series EHR data and achieve superior performance in the sepsis prediction tasks, so we use LSTM as the model backbone.
Note that the proposed active sensing methods are generalizable to various deep learning frameworks.

% \subsection{Implementation Details}

% We implement our proposed CAT-LSTM models with PyTorch 0.4.1\footnote{\url{https://pytorch.org/}}. For training models, we use Adam optimizer with a mini-batch of 256 patients.  We train on 4 GPUs (TITAN RTX 2080), with a learning rate of 0.0001. We randomly divide the patients in the dataset into 10 sets. All the experiment results are averaged from 10-fold cross validation, in which 7 sets are used for training every time, 1 set for validation and 2 sets for test. The validation sets are used to determine the best values of parameters in the training iterations.  Following \cite{kendall2017uncertainties,kendall2015bayesian}, we use 50 Monte Carlo dropout samples and use the average prediction as output for each patient. We use the area under the receiver operating characteristic curve (AUROC) in the test sets as a measure for comparing the performance of all the methods. More details can be found at  GitHub\footnote{\label{github}\url{https://github.com/anonymous13756/RAS}}.   

\section{Results}

\label{sec:result}
We now report the performance of \system in the three datasets.
We focus on answering the following research questions by our experimental results:
\begin{itemize}[leftmargin=*]
    \item \textbf{Q1: How does the model uncertainty affect the sepsis prediction performance?}
    % \item \textbf{Q2: Does the proposed uncertainty propagation method effectively and efficiently quantify the propagated uncertainty?}
    \item \textbf{Q2: How does the active sensing algorithm reduce the propagated uncertainty?}
    \item \textbf{Q3: How does the active sensing algorithm improve the sepsis prediction performance?}
    % \item \textbf{Q4: How to use the \system in real-world clinical scenarios?}
\end{itemize} 

\subsection{Q1: How does the model uncertainty affect the sepsis prediction performance?}

The existence of uncertainty makes AI models less reliable and less accurate when applying the models to real-world high-stakes scenarios.
% Moreover, high uncertainty might also make the model less accurate. 
In this subsection, we aim to show how the model uncertainty affects sepsis prediction performance by analyzing the relation between uncertainty and prediction performance. 

\subsubsection{Prediction Performance over Uncertainty Scales}
\begin{figure}
    \centering
    \vspace{3mm}
    \includegraphics[width=0.235\textwidth]{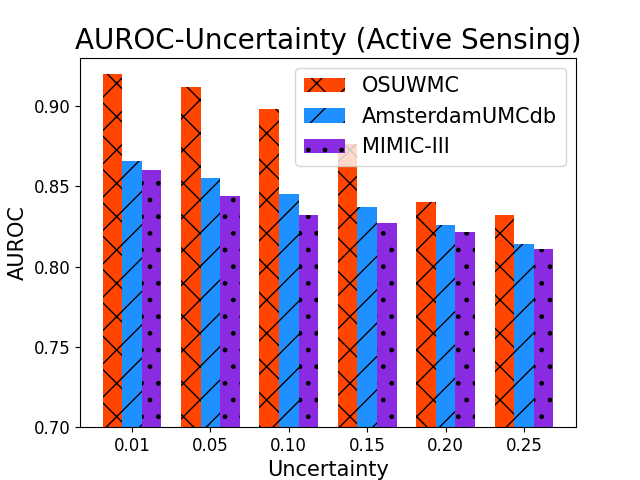}
    \includegraphics[width=0.235\textwidth]{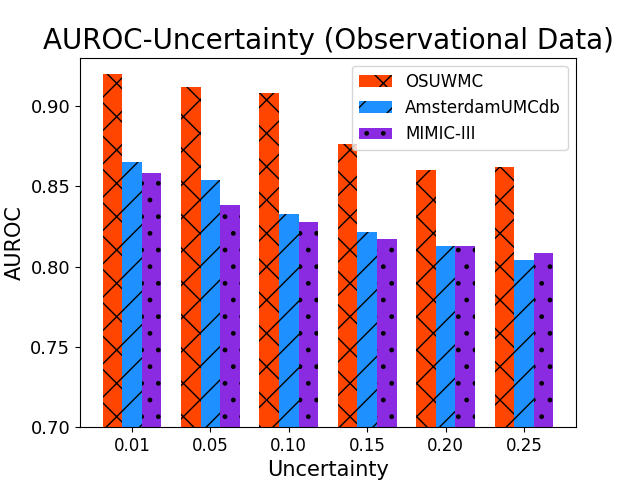}
    % \vspace{-3mm}
    \caption{Sepsis onset prediction performance with different uncertainty. }
    % \vspace{-3mm}
    \label{fig:uncertainty_auc}
\end{figure}

We compute the uncertainty of the sepsis onset prediction model's output with \autoref{eq:uncertainty_definition} and split the patients into 6 sets with different uncertainty scales. 
Then we calculate the sepsis onset prediction performance on AUROC inside each set.
\autoref{fig:uncertainty_auc} displays the model performance over the different uncertainty scales in the three datasets. 
We conducted experiments in two settings. In active sensing setting, we compute the AUROC after active sensing algorithms are used. In the observed data setting, we directly run the data in the observed data (including all the recorded vital signs and lab tests) and compute the AUROC. 
The results in both settings show that when uncertainty is higher, the model performance becomes less accurate, so a good active sensing framework can improve the prediction performance by reducing the uncertainty of the prediction model's output.

\subsubsection{Uncertainty Scales over Time}

\begin{figure}
    \centering
    \includegraphics[width=0.155\textwidth]{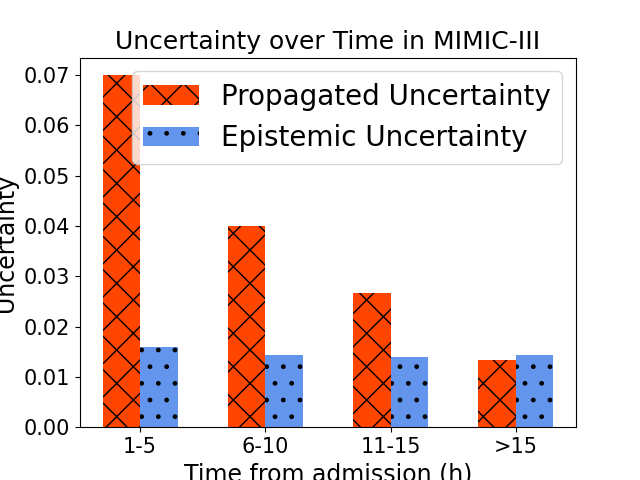} 
    \includegraphics[width=0.155\textwidth]{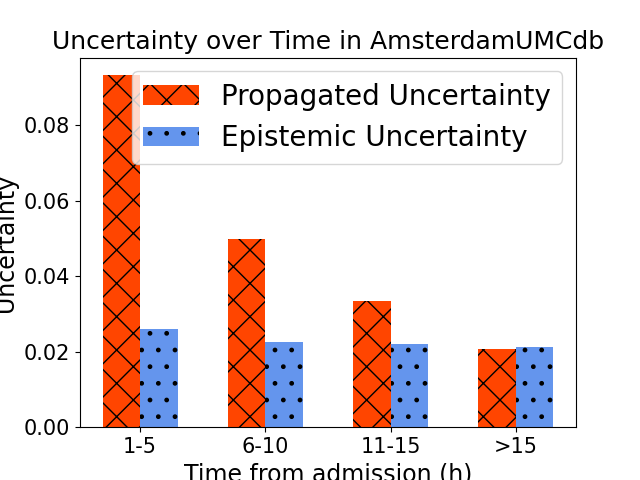} 
    \includegraphics[width=0.155\textwidth]{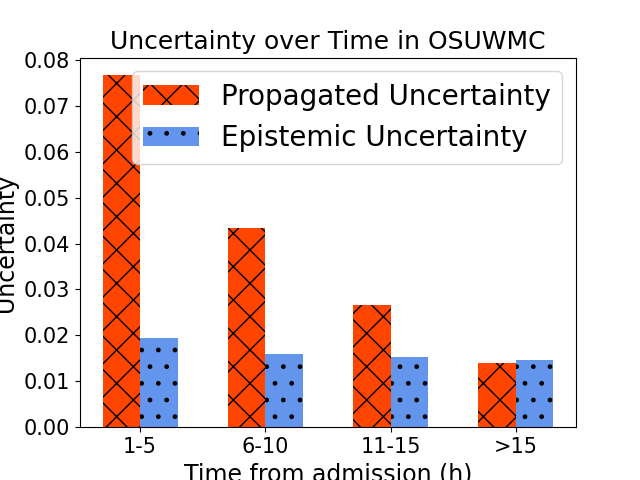} 
    % \vspace{-3mm}
    \caption{Uncertainty distribution over times after admission.} 
    % \vspace{-3mm}
    \label{fig:uncertainty_time}
\end{figure}

We quantify the model uncertainty at different times from admissions. 
\autoref{fig:uncertainty_time} displays the average uncertainty scales.

\autoref{fig:uncertainty_time} shows that in the first 15 hours, propagated uncertainty is dominant in sepsis onset risk prediction models. 
We speculate the reason is that at the beginning most variables have not been observed and the missing values cause the main uncertainty, which is consistent with our clinical experts' experience. With more variables collected, the propagated uncertainty decreases a lot after 15 hours of the admissions.

Because the missing variables can cause high uncertainty during the first hours, it is critical to quantify the propagated uncertainty when applying risk prediction models to high missing-rate settings. 
% Moreover,  it is also necessary to provide a UI interface to clinicians to illustrate whether the models are confident or uncertain about the prediction results.

\begin{figure}
    \centering
    \includegraphics[width=0.155\textwidth]{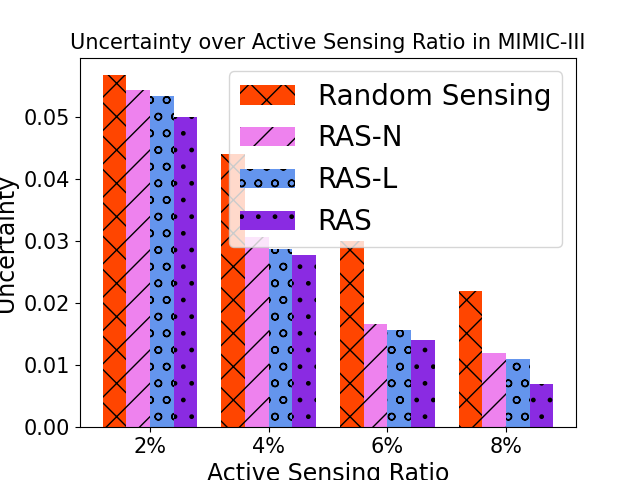} 
    \includegraphics[width=0.155\textwidth]{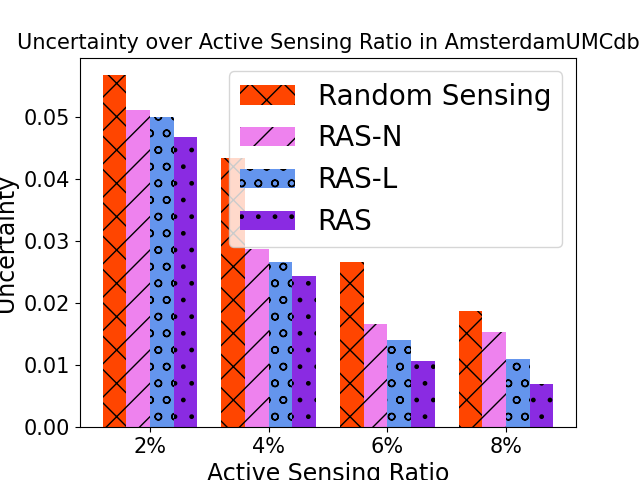} 
    \includegraphics[width=0.155\textwidth]{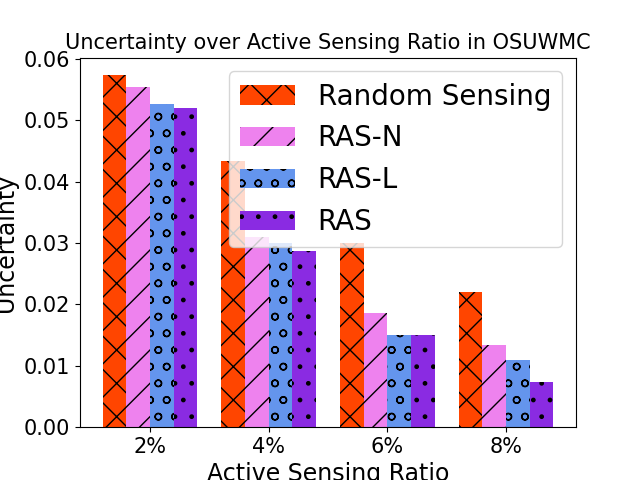} 
    % \vspace{-3mm}
    \caption{Uncertainty over different active sensing ratios.} 
    % \vspace{-3mm}
    \label{fig:uncertainty_active_sensing}
\end{figure}

% \subsection{Q2: Does the proposed uncertainty propagation method effectively and efficiently quantify the propagated uncertainty?}

% \CY{This subsection might be put into appendix.}

\subsection{Q2: How does the active sensing algorithm reduce the propagated uncertainty?}

Based on the estimated uncertainty, we propose active sensing algorithms to further reduce the prediction uncertainty by recommending clinicians collect more unobserved variables. We conduct experiments to show whether uncertainty can be significantly reduced with minimal additional variables observed. 

\subsubsection{Uncertainty with Different Active Sensing Ratio}

\autoref{fig:uncertainty_active_sensing} displays the average uncertainties for sepsis prediction results with different active sensing ratios.
The results show that with more missing variables observed, the uncertainty on the predicted sepsis risks are significantly reduced.
Besides, all the versions of the proposed RAS reduce more uncertainty than the baselines, 
which demonstrates the effectiveness of the proposed active sensing algorithms on uncertainty reduction.  

% \subsubsection{Propagated Uncertainty Quantification Accuracy} 
  
% We propose two uncertainty quantification methods. The first method is to sample the missing variables according to the data distribution and then compute uncertainty with \autoref{eq:uncertainty_definition}.  The second method is to use the imputation uncertainty times model gradient to estimate the propagated uncertainty. We use Monte-Carlo sampling to estimate the propagated uncertainty as the ground truth. Then we compare the estimated uncertainty with the ground truth to compute the estimation error. \autoref{tab:uncertainty_quantification_accuracy} displays the performance of uncertainty quantification. The results show that the introduced adversarial training can reduce more than 50\% of uncertainty estimation errors, which demonstrate the effectiveness of our uncertainty quantification method.

% \subsubsection{}
\subsubsection{Uncertainty Quantification Efficiency}
\begin{figure}
    \centering 
    \vspace{3mm}
    \includegraphics[width=0.155\textwidth]{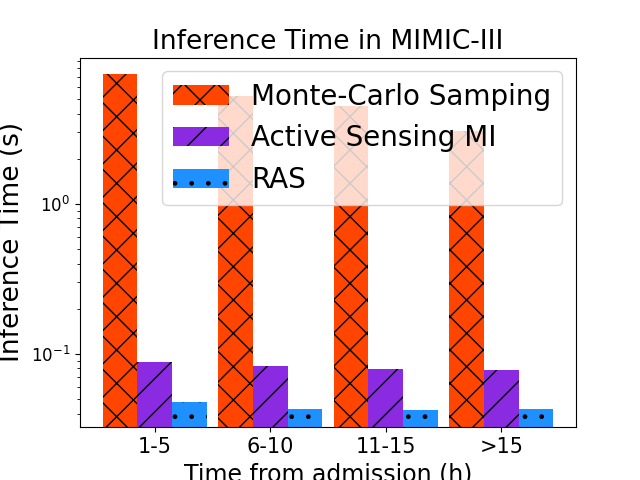} 
    \includegraphics[width=0.155\textwidth]{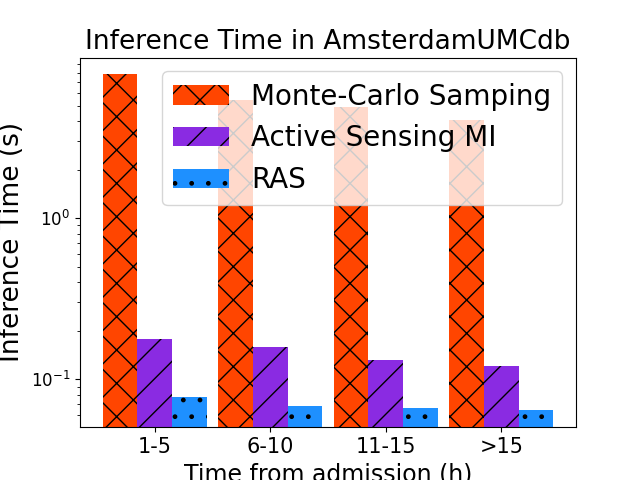} 
    \includegraphics[width=0.155\textwidth]{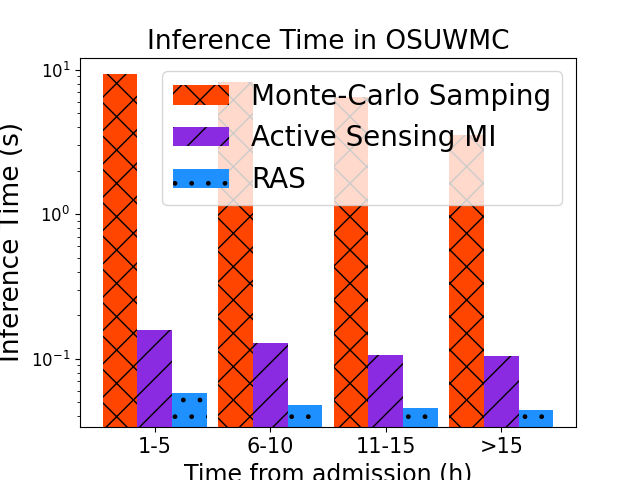}
    % \vspace{-3mm}
    \caption{Inference time cost over times after admission.} 
    % \vspace{-3mm}
    \label{fig:uncertainty_inference_time}
\end{figure}

We also investigate the time cost for uncertainty quantification during the inference phase.
\autoref{fig:uncertainty_inference_time} displays the inference time cost for uncertainty quantification. The results show that  RAS can achieve much less time than the baselines, which makes the \system system work more efficiently during the active sensing phase.

\begin{table*} [!t]
\centering
\caption{AUROC of risk prediction with the active sensing (cold-start).} 
    % \vspace{-3mm}
\label{tab:active_sensing_result_cold} 
\begin{tabular}{l|cccc|cccc|cccc}
\toprule 
 & 
\multicolumn{4}{c|}{MIMIC-III} & 
\multicolumn{4}{c|}{AmsterdamUMCdb} & 
\multicolumn{4}{c}{OSUWMC} \\
% & 0\% & 2\% & 4\% & 6\% & 8\% & All &0\% & 2\% & 4\% & 6\% & 8\% & All &0\% & 2\% & 4\% & 6\% & 8\%& All \\
&   2\% & 4\% & 6\% & 8\%  & 2\% & 4\% & 6\% & 8\% & 2\% & 4\% & 6\% & 8\%\\
\midrule
Random Sensing       & 0.761 & 0.772 & 0.779 & 0.785 & 0.772 & 0.781 & 0.788 & 0.793 & 0.785 & 0.794 & 0.805 & 0.811\\
Monte Carlo          & 0.771 & 0.789 & 0.797 & 0.812 & 0.782 & 0.795 & 0.802 & 0.817 & 0.797 & 0.818 & 0.855 & 0.886\\
Active Learning      & 0.773 & 0.791 & 0.804 & 0.817 &  0.780 &   0.800 & 0.805 & 0.816 & 0.802 &  0.820 & 0.857 & 0.889\\
% Active Sensing MI    & 0.776 & 0.797 & 0.809 & 0.819 & 0.784 & 0.801 & 0.814 & 0.817 & 0.802 & 0.835 & 0.867 & 0.914\\
VAT                  & 0.783 & 0.801 & 0.812 & 0.822 & 0.786 & 0.802 & 0.815 & 0.823 & 0.809 & 0.844 & 0.889 & 0.916\\
\midrule
RAS$^N$              &  0.770 & 0.788 & 0.796 &  0.810 &  0.780 & 0.793 & 0.802 & 0.814 & 0.795 & 0.816 & 0.853 & 0.881\\
RAS$^L$              & 0.783 & 0.801 & 0.812 & 0.824 & 0.785 & 0.801 & 0.818 & 0.822 & 0.814 & 0.848 & 0.877 & 0.917\\
RAS (Ours)                 & \textbf{0.792} & \textbf{ 0.810} & \textbf{0.823} & \textbf{0.835} & \textbf{0.795} & \textbf{0.809} & \textbf{0.828} & \textbf{ 0.840} & \textbf{0.823} & \textbf{0.857} & \textbf{0.889} & \textbf{0.929}\\
\bottomrule
\end{tabular}  
    % \vspace{-3mm}
\end{table*}

\subsection{Q3: How does the active sensing algorithm improve the sepsis prediction performance?}

The goal of \system is to accurately predict the sepsis so as to provide reliable decision-making support to clinicians. We conduct experiments to show sepsis prediction performance improvement with the active sensing algorithms.  

\subsubsection{Sepsis onset Prediction Results}
\autoref{tab:active_sensing_result_cold} displays the risk prediction performance with different active sensing ratios (i.e., 2\%-8\%). 
With additional variables observed, all the methods can achieve more accurate prediction performance for sepsis onset. 
Moreover, all the active sensing algorithms outperform the random sensing baseline with the same observation rate, which demonstrates that active sensing can improve downstream tasks' performance. 
Among the active sensing algorithms, the proposed RAS achieved the best performance with different active sensing ratios, which demonstrate the effectiveness of the proposed model. 
% Among the three versions of the proposed model, RAS outperforms RAS$^L$ and RAS$^N$, which demonstrates the effectiveness of the introduced adversarial training. 

\subsubsection{Ablation Study}

We have three versions of the framework. RAS$^N$ directly uses the gradient to estimate propagated uncertainty. RAS$^L$ uses a linear regularization term to make the model locally smooth, while RAS uses adversarial training. For RAS$^L$ and RAS versions, the additional terms change the loss functions. We conduct experiments to show whether the additional terms can improve model training.
We train the three versions of models independently and test them on all the observed data (without active sensing). 
As \autoref{tab:ablation_study} shows, RAS$^L$ and RAS outperform RAS$^N$, which demonstrates local linearity can further improve prediction performance. 
With adversarial training, RAS can achieve better local linearity than RAS$^L$ and thus perform the best, which also explains why the RAS outperforms better than Monte-Carlo sampling in \autoref{tab:active_sensing_result_cold} in the active sensing.

 We also conduct more experiments with different backbones (e.g., RNN, GRU, FC) and display the performance in \autoref{tab:backbones} in \autoref{sec:backbone}.  The experimental results show that the proposed model can consistently improve the prediction performance for all the backbones by recommending the most informative variables for observation.

% \autoref{tab:active_sensing_result_2} displays the improvements with different active sensing strategies. The results show that the improvements of RAS$^{AT}$ are better than the baselines by more than 20\% in both tasks, which demonstrate uncertainty propagation can effectively guide the active sensing algorithms to select the most informative variables when the learned target prediction functions are locally smooth.

\begin{table} [!t]
\centering
\caption{Sepsis prediction performance of three versions of RAS on observational data. } 
    % \vspace{-3mm}
\label{tab:ablation_study} 
\setlength{\tabcolsep}{1pt}
\begin{tabular}{lccc}
\toprule 
Method & 
MIMIC-III & 
AmsterdamUMCdb & OSUWMC \\
\midrule
RAS$^N$ &  0.820 & 0.820&  0.903\\
RAS$^L$ & 0.832 & 0.834 & 0.925\\
RAS &  \textbf{0.837} & \textbf{0.849}& \textbf{0.934}\\
\bottomrule
\end{tabular}  
    % \vspace{-3mm}
\end{table}

\subsubsection{Hyper-parameter Optimization}
The proposed RAS have four important hyper-parameter: weight $\alpha$ in \autoref{eq:weighted_loss}, step size $s_{adv}$, step $n_{adv}$, learning rate $lr$ in Algorithm \ref{alg:AT}. We use grid-search to find the best parameter (with active sensing ratio equal to 8\%). Table 5 displays the searching space and the optimal values used in the training process.

\begin{table}[]
    \centering
    \label{tab:hyper}
    \caption{The search space of hyper-parameters and optimal parameters utilized during the model training.}
    \begin{tabular}{lcc}
        \toprule
        Parameters & Search Space & Optimal Value \\
        \midrule
        Weight $\alpha$ & [0.1, 0.3, 0.5, 0.7, 0.9] & 0.5  \\
        Learning rate $lr$ & [1e-3, 1e-4, 1e-5] & 1e-4 \\
        Step size $s_{adv}$& [1e-2, 1e-3, 1e-4, 1e-5] & 1e-3 \\
        Step $n_{adv}$ & [1,2,5, 10, 15, 20] & 15 \\
    \bottomrule
    \end{tabular}
\end{table}

% \subsection{\system System}

\section{Deployment}

\label{sec:deployment}

\begin{figure}
    \centering
    \includegraphics[width=0.9\linewidth]{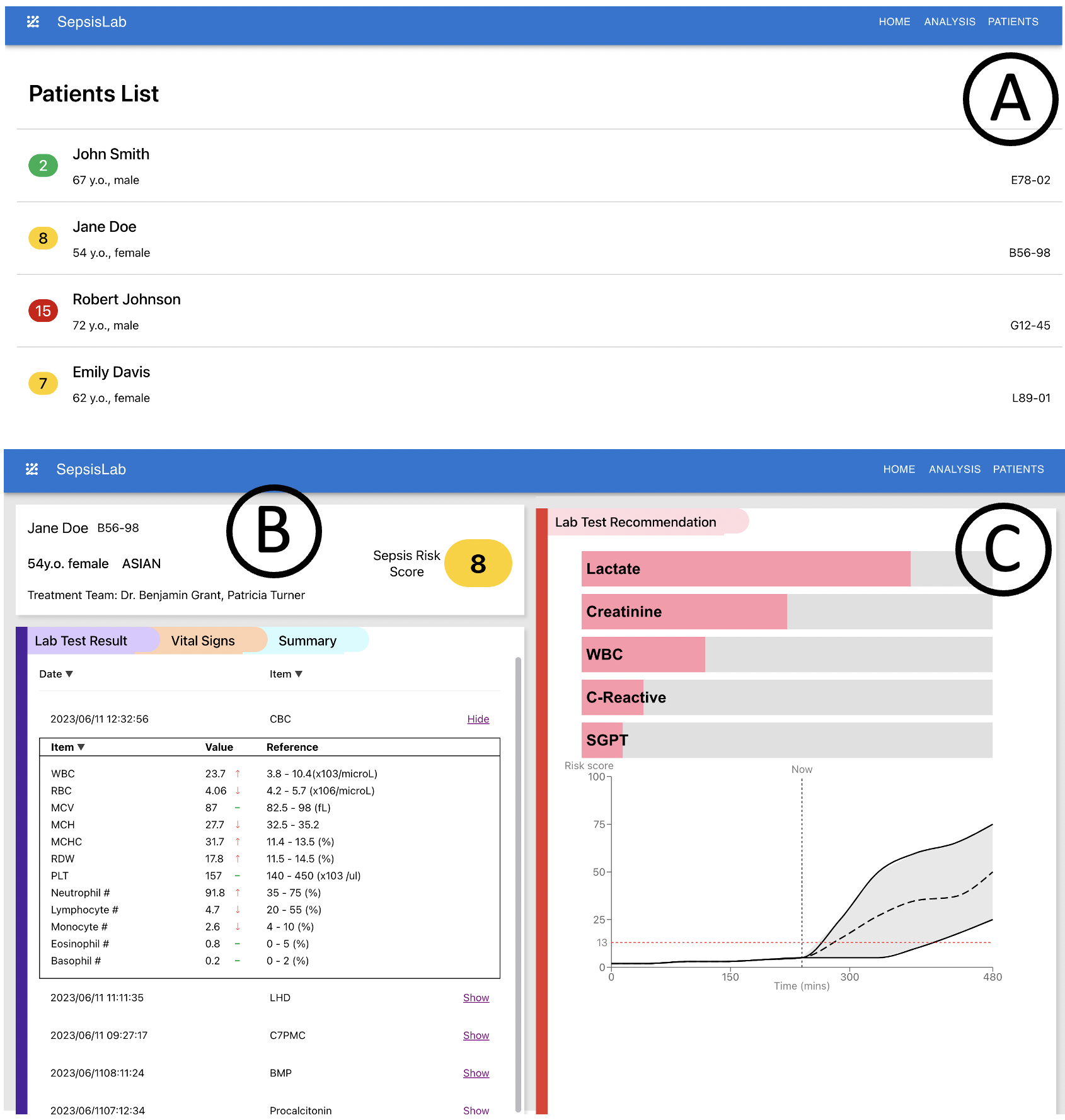}
    \caption{\textbf{User Interface of Our \system System. (A) Patient list with sepsis risk prediction score. (B) The patient’s demographics and the dashboard of the patient’s historical observations. (C) Predicted sepsis risk score with uncertainty range and recommended lab test items to observe.} }
    \label{fig:UI}
    \vspace{-2mm}
\end{figure}

\begin{figure}
    \centering
    \includegraphics[width=\linewidth]{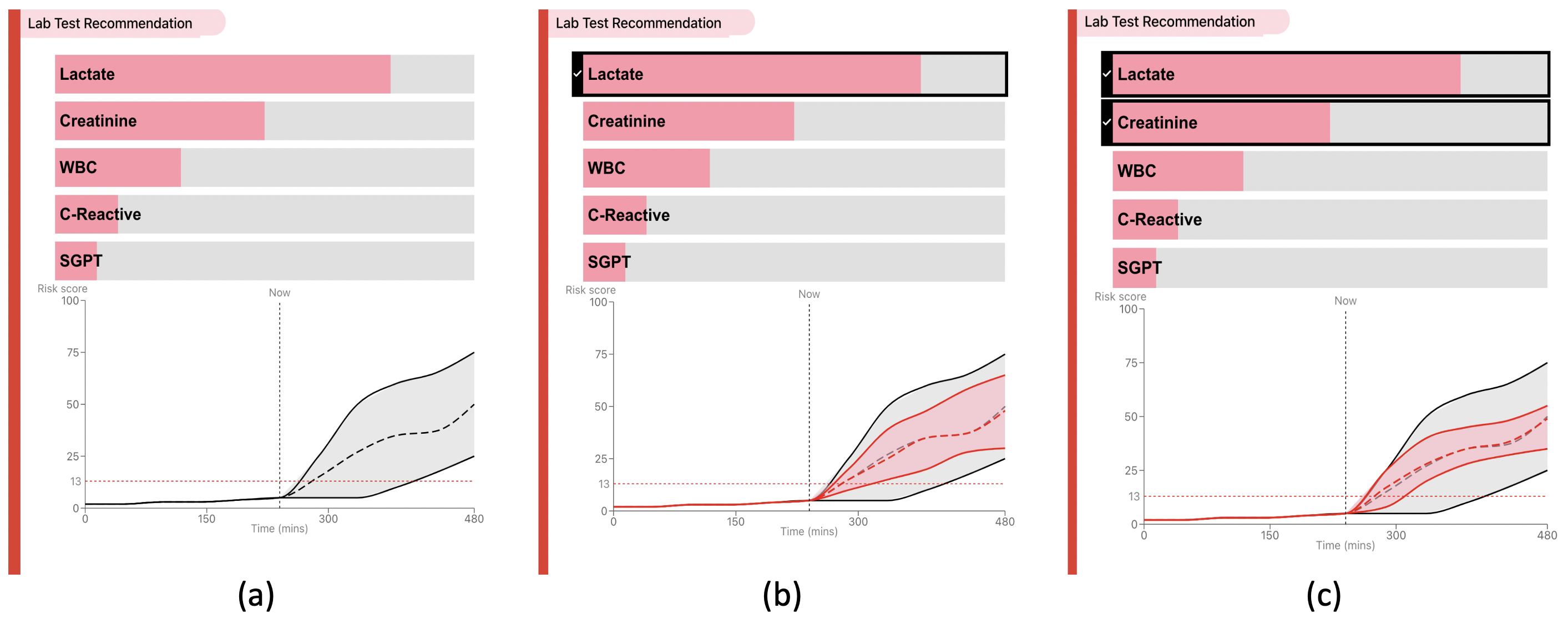}
    \caption{\textbf{The Interactive Lab Test Recommendation Module in \system System.}
    }
    \label{fig:UI_uncertainty}
\end{figure}

Based on the sepsis prediction model and active sensing algorithm, we implement a system \system.
\autoref{fig:UI} and \autoref{fig:UI_uncertainty} shows how the system is deploed in the Epic EHR Systems\footnote{\label{epic}\url{https://www.epic.com/software/}} at OSUWMC. 

\system starts to collect patients' data after the patients arrive hospital and automatically predicts sepsis risks hourly. 
\autoref{fig:UI}(A) displays a list of patients with different sepsis risk prediction scores, colored from no risk as Green, to medium risk as Yellow, to high risk as Red.
When picking a patient's data,
\autoref{fig:UI}(B) shows the patient's demographics and the dashboard that includes the patient's vital signs, lab test results, and medical history, which are helpful for clinicians to understand the patient's health states. 
\autoref{fig:UI}(C)  shows the patient’s sepsis risk (solid line) and uncertainty range (gray area) at different times
and an actionable lab item test recommendation list from \system. The items are ranked by their importance to reduce the uncertainty of the sepsis future
prediction.

The interactive process with our system is visualized in Figure~\ref{fig:UI_uncertainty}.
This UI currently illustrates that a clinical expert is examining a high-risk patient's data who was admitted 4 hours ago.
The \system suggests the expert collect more lab results. The expert is interacting with the visualization to see if Lactate and Creatinine lab results were added, and how the sepsis prediction and its uncertainty would change. 
The clinician can select a lab item (\autoref{fig:UI_uncertainty}(b)) or multiple lab items (\autoref{fig:UI_uncertainty}(c)) and see the expected influence of the lab test result on the model uncertainty via a counterfactual prediction. 
By comparing different combinations of the lab test items, the clinician can obtain
a better understanding of the model and make the decision to order appropriate lab tests to collect the actual item
values, which then truly update the model’s prediction trajectory and uncertainty range. 

Note that we used OSUWMC data for our algorithm illustration.
All patients' names and demographic info in this \autoref{fig:UI} are randomly generated for illustration purposes. 
Ongoing deployment also includes recruit clinicians for usability evaluation to quantitative and qualitatively measure clinical outcome and user satisfaction of SepsisLab (OSUWMC IRB\#: 2020H0018).

%% file: conclusion.tex
\section{Conclusion}

\label{sec:conclusion}
In this work, we study a real-world problem that how to accurately predict sepsis with limited variables available. 
Missing values widely exist in clinical data and can cause inaccurate prediction and high uncertainty for the sepsis prediction models.
To the best of our knowledge, it is the first work that studies the model uncertainty caused by missing values. 
We define a new term propagated uncertainty to describe the uncertainty, which is the downstream models' uncertainty propagated from the uncertain input (i.e., imputation results).
We further propose uncertainty propagation methods to quantify the propagated uncertainty.
Based on the uncertainty quantification, we propose a robust active sensing algorithm to reduce the uncertainty by actively recommending clinicians to observe the most informative variables. 
The experimental results on real-world datasets show that the introduced propagated uncertainty is dominant at the beginning of patients' admissions to the hospital due to the very limited variables and the proposed active sensing algorithm can significantly reduce the propagated uncertainty and thus improve the sepsis prediction performance. 
Finally, we design a \system system for deployment to integrate into clinicians' workflow, which paves the way for human-AI collaboration and early intervention for sepsis management.

%% file: acknowledgement.tex
\section{ACKNOWLEDGMENTS}
    This work was funded in part by the National Science Foundation under award number IIS-2145625, by the National Institutes of Health under award number R01GM141279 and R01AI188576, and by The Ohio State University President’s Research Excellence Accelerator Grant.

%% file: appendix.tex
\appendix
 
\section{Appendix}

\begin{table*} [ht]
\centering
\caption{AUROC of risk prediction of the propose RAS with different backbones.} 
    % \vspace{-3mm}
\label{tab:backbones}  
\begin{tabular}{l|cccc|cccc|cccc}
\toprule 
 & 
\multicolumn{4}{c|}{MIMIC-III} & 
\multicolumn{4}{c|}{AmsterdamUMCdb} & 
\multicolumn{4}{c}{OSUWMC} \\ 
&   2\% & 4\% & 6\% & 8\%  & 2\% & 4\% & 6\% & 8\% & 2\% & 4\% & 6\% & 8\%\\
\midrule
FC              &  0.760 & 0.772 & 0.782 &  0.792 &  0.772 & 0.785 & 0.792 & 0.801 & 0.785 & 0.801 & 0.823 & 0.841\\
RNN              &  0.782 & 0.790 & 0.810 &  0.821 &  0.787 & 0.799 & 0.816 & 0.825 & 0.814 & 0.842 & 0.873 & 0.910\\
GRU              & 0.789 & 0.799 & 0.822 & 0.833 & 0.792 & 0.806 & 0.826 & 0.837 & 0.821 & 0.855 & 0.887 & 0.928\\
LSTM                 & \textbf{0.792} & \textbf{ 0.81} & \textbf{0.823} & \textbf{0.835} & \textbf{0.795} & \textbf{0.809} & \textbf{0.828} & \textbf{ 0.840} & \textbf{0.823} & \textbf{0.857} & \textbf{0.889} & \textbf{0.929}\\
\bottomrule
\end{tabular}   
\end{table*}

\subsection{Uncertainty Propagation for Linear Function}
\label{sec:linear_uncertainty}

For a linear function $f_w(x) = \sum_{i} w_i x_i$ ($1 \le i \le n$), the uncertainty is defined as the variance: 
$$Var(f_w(x)) = \int_x (f_w(x) - f_w(\mu_x))^2 dx$$
$$ = \int_x (\sum_{i} w_i x_i - \sum_{i} w_i \mu_i)^2dx$$  
$$ = \int_x [\sum_{i} (w_i x_i - w_i \mu_i)]^2dx$$ 
$$ = \int_x \sum_{i} (w_i x_i - w_i \mu_i)\sum_{j} (w_j x_j - w_j \mu_j)dx$$  
$$ = \int_x \sum_{i} \sum_{j} (w_i x_i - w_i \mu_i) (w_j x_j - w_j \mu_j)dx$$   
$$ = \int_x \sum_{i} \sum_{j=i} (w_i x_i - w_i \mu_i) (w_i x_i - w_i \mu_i) $$
$$\qquad + \sum_{i} \sum_{j\neq i} (w_i x_i - w_i \mu_i) (w_j x_j - w_j \mu_j)dx$$
$$=\int_x \sum_i w_i^2 \sigma_i^2 + \sum_i \sum_{j\neq i} w_i w_j \rho_{i,j} \sigma_i \sigma_j dx$$
where $\mu_x$ denotes the mean of variable $x$. $i$ and $j$ denote the indices of variables or parameters. $\rho$ denotes the correlation coefficient. $\sigma$ denotes the standard deviation. 

\subsection{Missing Rates of Clinical Variables}
\label{sec:missing-rate}

We display the missing rates of lab test variables in \autoref{tab:all-variables}.

\begin{table} [!htb] 
% \vspace{-2mm}
\setlength{\tabcolsep}{2.2pt}
\centering
\caption{Missing rates of observed lab tests.}  
\label{tab:all-variables} 
\begin{tabular}{cccc}
\toprule
variable & AmsterdamUMCdb & OSUWMC & MIMIC-III\\
\midrule
WBC                & 67\% & 78\% & 69\%\\
BUN                & 63\% & 76\% & 66\%\\
GCS                & 29\% & 50\% & 33\%\\
Urineoutput        & 23\% & 39\% & 33\%\\
Creatinine (CRT)   & 75\% & 85\% & 80\%\\
Platelet (PLT)     & 76\% & 88\% & 82\%\\
\midrule
Glucose (GLC)      & 34\% & 49\% & 36\%\\
Sodium (SDM)       & 55\% & 72\% & 65\%\\
Hemoglobin (HMG)   & 56\% & 75\% & 69\%\\
Chloride (CLR)     & 62\% & 70\% & 66\%\\
Bicarbonate (BCB)  & 69\% & 74\% & 67\%\\
Lactate (LCT)      & 88\% & 90\% & 89\%\\
\midrule
INR                & 78\% & 84\% & 80\%\\
PTT                & 76\% & 83\% & 79\%\\
Magnesium          & 66\% & 76\% & 69\%\\
Aniongap (AG)      & 62\% & 78\% & 67\%\\
Hematocrit (HMT)   & 60\% & 76\% & 64\%\\
PT                 & 78\% & 92\% & 80\%\\
\bottomrule
\end{tabular}  
\end{table}

\subsection{Model Performance with different backbones}
\label{sec:backbone}

Our model is applicable to various models, including LSTM, GRU, and fully-connected networks (FC).
LSTM has shown superior performance in modeling clinical time series data in multiple tasks, including missing value imputation \cite{yin2019domain,yan2019deep}, clinical prediction \cite{men2021multi}, and patient subtyping \cite{baytas2017patient}, so we choose LSTM as the model backbone. We also conducted more experiments with different backbones as shown in \autoref{tab:backbones}. The experimental results show that the proposed model can significantly improve the prediction performance for all the backbones by recommending the most informative variables for observation.